\documentclass[10pt,twocolumn,letterpaper]{article}

\usepackage{iccv}
\usepackage{times}
\usepackage{epsfig}
\usepackage{graphicx}
\usepackage{amsmath}
\usepackage{amssymb}
\usepackage{adjustbox,subfigure}
\usepackage{enumitem}

\usepackage[numbers,sort]{natbib} % sort citations

\iccvfinalcopy % *** Uncomment this line for the final submission

\def\iccvPaperID{****} % *** Enter the ICCV Paper ID here

\begin{document}

%%%%%%%%% TITLE
\title{Multi-task Self-Supervised Visual Learning}

\author{Carl  Doersch$^\dagger$
\and
Andrew Zisserman$^{\dagger,*}$
\and
$^\dagger$DeepMind
\quad\quad\quad
$^*$VGG, Department of Engineering Science, University of Oxford\\
}

\maketitle
%\thispagestyle{empty}

%%%%%%%%% ABSTRACT
\begin{abstract}
We investigate methods for combining multiple self-supervised
tasks---i.e., supervised tasks where data can be collected without
manual labeling---in order to train a single visual representation.
First, we provide an apples-to-apples comparison of four different
self-supervised tasks using the very deep ResNet-101 architecture.  
We then combine tasks to jointly train a network. We also explore lasso regularization to encourage the
network to factorize the information in its representation, and methods for ``harmonizing'' network inputs in order
to learn a more unified representation.  
We evaluate all methods on ImageNet classification, PASCAL VOC detection, and NYU depth prediction.
Our results show that deeper networks work better, and that combining tasks---even via a na{\"i}ve multi-head architecture---always improves performance.
Our best joint network nearly matches 
the PASCAL performance of a model pre-trained on ImageNet classification, and matches the ImageNet network on NYU depth prediction.
\end{abstract}

%%%%%%%%% BODY TEXT

\section{Introduction}

Vision is one of the most promising domains for unsupervised learning.
Unlabeled images and video are available in practically unlimited quantities, and the most prominent present image models---neural networks---are data starved, easily memorizing even random labels for large image collections~\cite{zhang2016understanding}.
Yet unsupervised algorithms are still not very effective for training neural networks:
they fail to adequately capture the visual semantics needed to solve real-world tasks like object detection or geometry estimation the way strongly-supervised methods do.  
For most vision problems, the current state-of-the-art approach begins by training a neural network on ImageNet~\cite{russakovsky2015imagenet} or a similarly large dataset which has been hand-annotated.

How might we better train neural networks without manual labeling?
Neural networks are generally trained via backpropagation on some objective function.
Without labels, however, what objective function can measure how good the network is?
\textit{Self-supervised} learning answers this question by proposing various tasks for networks to solve, where performance is easy to measure, i.e., performance can be captured with an objective function like those seen in supervised learning.
Ideally, these tasks will be difficult to solve without understanding some form of image semantics, yet any labels necessary to formulate the objective function can be obtained automatically.
In the last few years, a considerable number of such tasks have been proposed~\cite{agrawal2015learning,agrawal2016learning,denton2016semi,doersch2015unsupervised,dosovitskiy2014discriminative,jayaraman2015learning,larsson2016learning,li2016unsupervised,misra2016shuffle,noroozi2016unsupervised,owens2016ambient,pathak2016learning,pathak2016context,pinto2016supervision,pinto2016supersizing,walker2016uncertain,walker2015dense,wang2015unsupervised,wiskott2002slow,zhang2016colorful,zou2012deep}, such as asking a neural network to colorize grayscale images, fill in image holes, solve jigsaw puzzles made from image patches, or predict movement in videos.
Neural networks pre-trained with these tasks can be re-trained to perform well on standard vision tasks (e.g.\ image classification, object detection, geometry estimation) with less manually-labeled data than networks which are initialized randomly.  
However, they still perform worse in this setting than networks pre-trained on ImageNet.

This paper advances self-supervision first by implementing four self-supervision tasks and comparing their performance using three evaluation measures. 
The self-supervised tasks are: relative position~\cite{doersch2015unsupervised}, colorization~\cite{zhang2016colorful}, the ``exemplar" task~\cite{dosovitskiy2014discriminative}, and motion segmentation~\cite{pathak2016learning} (described in section~\ref{sec:tasks}). 
The evaluation measures (section~\ref{sec:evaluation}) assess a diverse set of applications that are standard for this area, including ImageNet image classification, object category detection on PASCAL VOC 2007, and depth prediction on NYU v2.

Second, we evaluate if performance can be boosted by combining these
tasks to simultaneously train a single trunk network.  Combining the
tasks fairly in a multi-task learning objective is challenging since
the tasks learn at different rates, and we discuss how we handle this
problem in section~\ref{sec:training}.
We find that multiple tasks work better than one, and explore which
combinations give the largest boost.

Third, we identify two reasons why a na{\"i}ve combination of self-supervision tasks might conflict, impeding performance: \textit{input channels} can conflict, and \textit{learning tasks} can conflict.
The first sort of conflict might occur when jointly training colorization and exemplar learning: colorization receives grayscale images as input, while exemplar learning receives all color channels.
This puts an unnecessary burden on low-level feature detectors that must operate across domains.
The second sort of conflict might happen when one task learns semantic categorization (i.e. generalizing across instances of a class) and another learns instance matching (which should not generalize within a class).
We resolve the first conflict via ``input harmonization'', i.e.\ modifying network inputs so different tasks get more similar inputs.  
For the second conflict, we extend our mutli-task learning architecture with a lasso-regularized combination of features from different layers, which encourages the network to separate features that are useful for different tasks.
These architectures are described in section~\ref{sec:architectures}.

We use a common deep network across all experiments, a ResNet-101-v2, so that we can compare various diverse self-supervision tasks apples-to-apples.
This comparison is the first of its kind.  Previous work applied
self-supervision tasks over a variety of CNN architectures (usually
relatively shallow), and often evaluated the representations on
different tasks; and even where the evaluation tasks are the same, there
are often differences in the fine-tuning algorithms. Consequently, it has not been possible to 
compare the performance of different self-supervision tasks across papers.
Carrying out multiple fair comparisons, together with the implementation
of the self-supervised tasks, joint training, evaluations, and
optimization of a large network for several large datasets has been a
significant engineering challenge. We describe how we carried out the
large scale training efficiently in a distributed manner
in section~\ref{sec:training}. This is another contribution of the paper.

As shown in the experiments of section~\ref{sec:results}, by combining multiple
self-supervision tasks we are able to close further the gap between self-supervised and fully supervised 
pre-training over all three evaluation measures.

\subsection{Related Work}

Self-supervision tasks for deep learning 
generally involve taking a complex signal, hiding part of it from the network, and then asking the network to fill in the missing information.
The tasks can broadly be divided into those that use auxiliary information or those that only use raw pixels.

Tasks that use auxiliary information such as  multi-modal information beyond
pixels include:  predicting sound given
videos~\cite{owens2016ambient}, predicting camera motion given two
images of the same
scene~\cite{jayaraman2015learning,agrawal2015learning,zamir2016generic},
or predicting what robotic motion caused a change in a
scene~\cite{pinto2016curious,pinto2016supersizing,pinto2016supervision,agrawal2016learning,pinto2016learning}.
However, non-visual information can be difficult to obtain: estimating
motion requires IMU measurements, running robots is still expensive,
and sound is complex and difficult to evaluate
quantitatively.

Thus, many works use raw pixels.  
In videos, time can be a source of supervision.
One can simply predict future~\cite{walker2016uncertain,walker2015dense}, although such predictions may be difficult to evaluate.
One way to simplify the problem is to ask a network to temporally order a set of frames sampled from a video~\cite{misra2016shuffle}.
Another is to note that objects generally appear across many frames: thus, we can train features to remain invariant as a video progresses~\cite{foldiak1991learning,wiskott2002slow,mobahi2009deep,wang2015unsupervised,zou2012deep}. 
Finally, motion cues can separate foreground objects from background.
Neural networks can be asked to re-produce these motion-based boundaries without seeing motion~\cite{pathak2016learning,li2016unsupervised}.

Self-supervised learning can also work with a single image.  
One can hide a part of the image and ask the network to make predictions about the hidden part.
The network can be tasked with generating pixels, either by filling in holes~\cite{pathak2016context,denton2016semi}, or recovering color after images have been converted to grayscale~\cite{zhang2016colorful,larsson2016learning}.
Again, evaluating the quality of generated pixels is difficult.
To simplify the task, one can extract multiple patches at random from an image, and then ask the network to position the patches relative to each other~\cite{doersch2015unsupervised,noroozi2016unsupervised}.
Finally, one can form a surrogate ``class'' by taking a single image and altering it many times via translations, rotations, and color shifts~\cite{dosovitskiy2014discriminative}, to create a synthetic categorization problem.

Our work is also related to multi-task learning.
Several recent works have trained deep visual representations using multiple tasks~\cite{sermanet2013overfeat,eigen2015predicting,gkioxari2015contextual,misra2016cross}, including one work~\cite{kokkinos2016ubernet} which combines no less than 7 tasks.
Usually the goal is to create a single representation that works well for every task, and perhaps share knowledge between tasks.
Surprisingly, however, previous work has shown little transfer between diverse tasks.
Kokkinos~\cite{kokkinos2016ubernet}, for example, found a slight dip in performance with 7 tasks versus 2.  
Note that our work is not primarily 
concerned with the performance on the self-supervised tasks we combine: we evaluate on a separate set of semantic ``evaluation tasks.''  
Some previous self-supervised learning literature has suggested performance gains from combining self-supervised tasks~\cite{pinto2016learning,zamir2016generic}, although these works used relatively similar tasks within relatively restricted domains where extra information was provided besides pixels.  
In this work, we find that pre-training on multiple diverse self-supervised tasks using only pixels yields strong performance.

\section{Self-Supervised Tasks}

\label{sec:tasks}

Too many self-supervised tasks have been proposed in recent years for us to evaluate every possible combination.  
Hence, we chose representative self-supervised tasks to reimplement and investigate in combination.  
We aimed for tasks that were conceptually simple, yet also as diverse as possible.
Intuitively, a diverse set of tasks should lead to a diverse set of features, which will therefore be more likely to span the space of features needed for general semantic image understanding.
In this section, we will briefly describe the four tasks we investigated.
Where possible, we followed the procedures established in previous works, although in many cases modifications were necessary for our multi-task setup.

\paragraph{Relative Position~\cite{doersch2015unsupervised}:} This task begins by sampling two patches at random from a single image and feeding them both to the network without context.  
The network's goal is to predict where one patch was relative to the
other in the original image.  The trunk is used to produce a
representation separately for both patches, which are then fed into a
head which combines the representations and makes a prediction.
The patch locations are sampled from a grid, and pairs are always
taken from adjacent grid points (including diagonals).  Thus, there
are only eight possible relative positions for a pair, meaning the
network output is a simple eight-way softmax classification.
Importantly, networks can learn to detect chromatic aberration to
solve the task, a low-level image property that isn't relevant to
semantic tasks.  Hence,~\cite{doersch2015unsupervised} employs ``color
dropping'', i.e., randomly dropping 2 of the 3 color channels and
replacing them with noise.  We reproduce color dropping, though our
harmonization experiments explore other approaches to dealing with
chromatic aberration that clash less with other tasks.

\paragraph{Colorization~\cite{zhang2016colorful}:} Given a grayscale image (the L channel of the Lab color space), the network must predict the color at every pixel (specifically, the ab components of Lab).
The color is predicted at a lower resolution than the image (a stride
of 8 in our case, a stride of 4 was used in~\cite{zhang2016colorful}), and furthermore, the colors are vector quantized
into 313 different categories.  Thus, there is a 313-way softmax
classification for every 8-by-8 pixel region of the image.  Our
implementation closely follows~\cite{zhang2016colorful}.

\paragraph{Exemplar~\cite{dosovitskiy2014discriminative}:} The original implementation of this task created pseudo-classes, where each class was generated by taking a patch from a single image and augmenting it via translation, rotation, scaling, and color shifts~\cite{dosovitskiy2014discriminative}.  
The network was trained to discriminate between pseudo-classes.
Unfortunately, this approach isn't scalable to large datasets, since
the number of categories (and therefore, the number of parameters in the final fully-connected layer)
scales linearly in the number of images.  
However, the approach can be extended to
allow an infinite number of classes by using a triplet loss,
similar to~\cite{wang2015unsupervised}, instead of a classification loss
per class.  Specifically, we randomly
sample two patches $x_1$ and $x_2$ from the same pseudo-class, and a third patch $x_3$ from a different pseudo-class (i.e. from a different image).  The network is trained with a loss of the form
$max(D(f(x_1),f(x_2))-D(f(x_1),f(x_3))+M,0)$, where $D$ is the cosine
distance, $f(x)$ is network features for $x$ (including a
small head) for patch $x$, and $M$ is a margin which we set to~$0.5$.

\paragraph{Motion Segmentation~\cite{pathak2016learning}:} Given a single frame of video, this task asks the network to classify which pixels will move in subsequent frames.
The ``ground truth'' mask of moving pixels is extracted using standard dense tracking algorithms.
We follow Pathak {\it et al.}~\cite{pathak2016learning}, except that we replace their tracking algorithm with Improved Dense Trajectories~\cite{wang2013action}.  
Keypoints are tracked over 10 frames, and any pixel not labeled as camera motion by that algorithm is treated as foreground.  
The label image is downsampled by a factor of 8.
The resulting segmentations look qualitatively similar to those given in Pathak {\it et al.}~\cite{pathak2016learning}.
The network is trained via a per-pixel cross-entropy with the label image.

\paragraph{Datasets:} The three image-based tasks are all trained on ImageNet, as is common
in prior work.  The motion segmentation task uses the SoundNet dataset~\cite{aytar2016soundnet}. 
It is an open problem whether performance
can be improved by different choices of dataset, or indeed by training
on much larger datasets.

\section{Architectures}
\label{sec:architectures}

\begin{figure*}[t]
\begin{center}
%\begin{tabular}{l}

   \mbox{}%
   \adjustbox{valign=M}{a)\hspace{.5cm}}\adjincludegraphics[valign=M,width=0.85\linewidth]{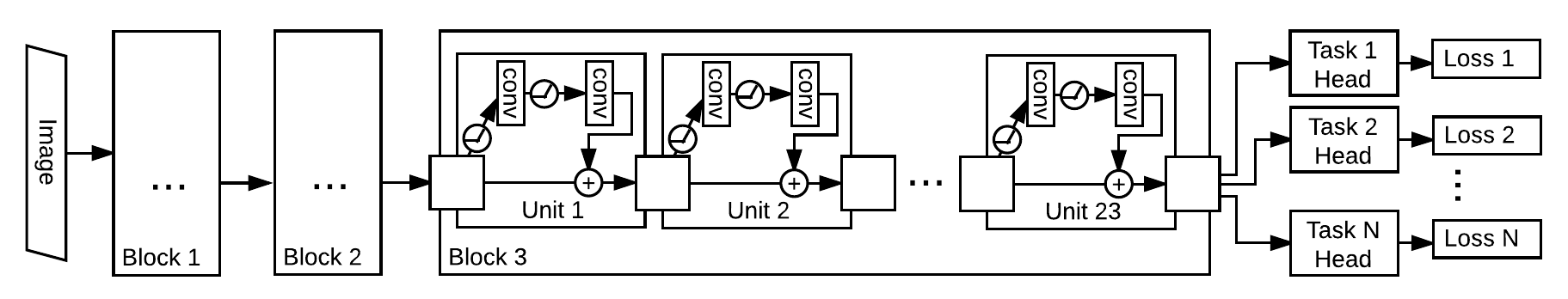}\\
   
   \adjustbox{valign=M}{b)\hspace{.5cm}}\adjincludegraphics[valign=M,width=0.85\linewidth]{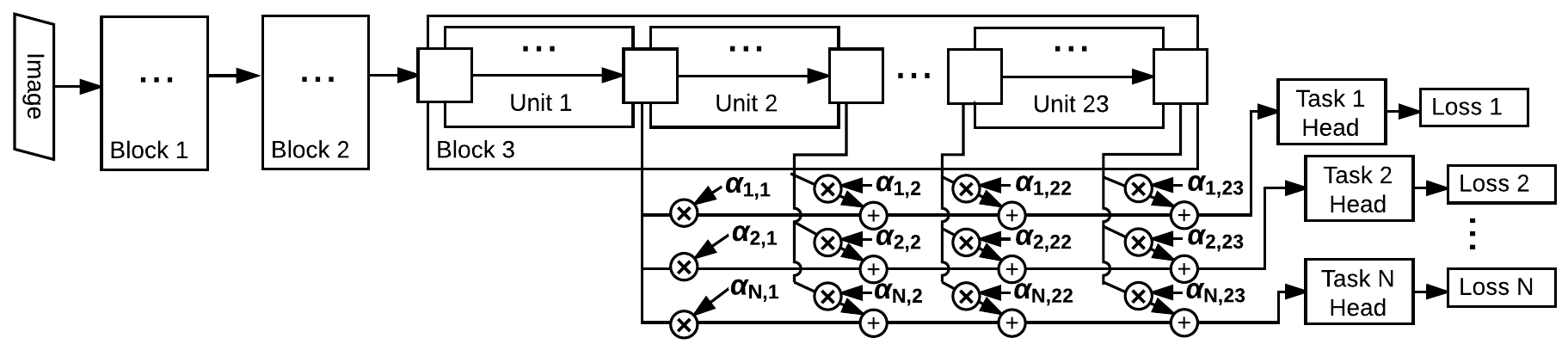}
%\end{tabular}
\end{center}

   \caption{
   The structure of our multi-task network. It is based on ResNet-101, with block 3 having 23 residual units.  a) Naive shared-trunk approach, where each ``head'' is attached to the output of block 3.  b) the lasso architecture, where each ``head'' receives a linear combination of unit outputs within block3, weighted by the matrix $\alpha$, which is trained to be sparse.
   }

\label{fig:net_structure}
\end{figure*}

In this section we describe three architectures: first, the (na{\"i}ve) multi-task network that has a common trunk and a head for each task (figure \ref{fig:net_structure}a); second, the lasso extension of this architecture (figure \ref{fig:net_structure}b) that enables the training to determine the combination of layers to use for each self-supervised task; and third, a method for harmonizing input channels across self-supervision tasks. 

\subsection{Common Trunk}

Our architecture begins with Resnet-101 v2~\cite{he2016identity}, as implemented in
TensorFlow-Slim~\cite{Guadarrama16}.  
We keep the entire architecture up to the end of block 3, and use the same block3
representation solve all tasks and
evaluations (see figure~\ref{fig:net_structure}a).  Thus, our 
``trunk'' has an output with 1024 channels, and consists of 88 convolution layers with roughly 30 million
parameters.  Block 4 contains an additional 13 conv layers and 20
million parameters, but we don't use it to save computation.

Each task has a separate loss, and has extra layers in a ``head,'' which may have a complicated structure.
For instance, the relative position and exemplar tasks have a siamese architecture.
We implement this by passing all patches through the trunk as a single batch, and then re-arranging the elements in the batch to make pairs (or triplets) of representations to be processed by the head.
At each training iteration, only one of the heads is active.
However, gradients are averaged across many iterations where different heads are active, meaning that the overall loss is a sum of the losses of different tasks.

\subsection{Separating features via Lasso}
\label{sec:lasso}

Different tasks require different features; this applies for both
the self-supervised training tasks and the evaluation tasks.  
For example, information
about fine-grained breeds of dogs is useful for, e.g., ImageNet
classification, and also colorization.  However, fine-grained
information is less useful for tasks like PASCAL object detection, or
for relative positioning of patches.  Furthermore, some tasks require
only image patches (such as relative positioning) whilst others can
make use of entire images (such as colorization), and consequently
features may be learnt at different scales.  This suggests that, while
training on self-supervised tasks, it might be advantageous to
separate out groups of features that are useful for some tasks but not
others.  This would help us with evaluation tasks: we expect that any
given evaluation task will be more similar to some self-supervised
tasks than to others.  Thus, if the features are factorized into
different tasks, then the network can select from the discovered
feature groups while training on the evaluation tasks.

Inspired by recent works that extract information across network layers for the sake of transfer learning~\cite{misra2016cross,hariharan2015hypercolumns,rusu2016progressive}, we
propose a mechanism which allows a network to choose which layers
are fed into each task.  The simplest approach might be to use a
task-specific skip layer which selects a single layer in ResNet-101
(out of a set of equal-sized candidate layers) and feeds it directly
into the task's head.  However, a hard selection operation isn't
differentiable, meaning that the network couldn't learn which layer to
feed into a task.  Furthermore, some tasks might need information from
multiple layers.  Hence, we relax the hard selection process, and instead pass
a linear combination of skip
layers to each head.  Concretely, each task has a set of coefficients, one for
each of the 23 candidate layers in block 3.  The representation
that's fed into each task head is a sum of the layer activations
weighted by these task-specific coefficients.  We impose a lasso (L1)
penalty to encourage the combination to be sparse, which therefore
encourages the network to concentrate all of the information required
by a single task into a small number of layers.  Thus, when
fine-tuning on a new task, these task-specific layers can be quickly
selected or rejected as a group, using the same lasso penalty.

Mathematically, we create a matrix $\alpha$ with $N$ rows and $M$ columns, where $N$ is the number of self-supervised tasks, and $M$ is the number of residual units in block 3.  
The representation passed to the head for task $n$ is then:

\begin{equation}
    \sum_{m=1}^{M}\alpha_{n,m}*Unit_m
\end{equation}

where $Unit_m$ is the output of residual unit $m$.
We enforce that $\sum_{m=1}^{M}\alpha_{n,m}^{2}=1$ for all tasks $n$, to control the output variance (note that the entries in $\alpha$ can be negative, so a simple sum is insufficient).
To ensure sparsity, we add an L1 penalty on the entries of $\alpha$ to the objective 
function.
We create a similar $\alpha$ matrix for the set of evaluation tasks.

\subsection{Harmonizing network inputs}

Each self-supervised task pre-processes its data differently, so the low-level image statistics are often very different across tasks.
This puts a heavy burden on the trunk network, since its features must generalize across these statistical differences, which may impede learning.
Furthermore, it gives the network an opportunity to cheat: the network might recognize which task it must solve, and only represent information which is relevant to that task, instead of truly multi-task features.
This problem is especially bad for relative position, which pre-processes its input data by discarding 2 of the 3 color channels, selected at random, and replacing them with noise.
Chromatic aberration is also hard to detect in grayscale images.
Hence, to ``harmonize,'' we replace relative position's preprocessing with the same preprocessing used for colorization: images are converted to Lab, and the a and b channels are discarded (we replicate the L channel 3 times so that the network can be evaluated on color images).

\begin{figure}[t]
\begin{center}
   \includegraphics[width=0.98\linewidth]{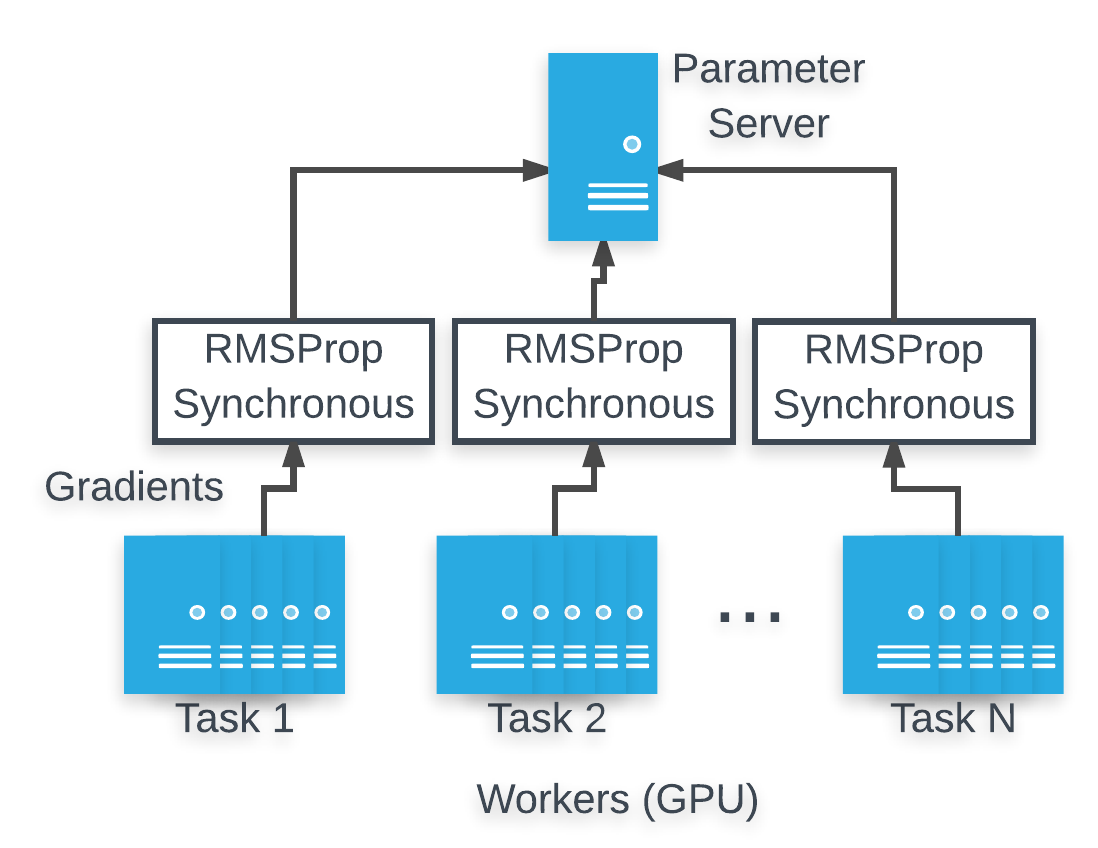}
\end{center}

   \caption{
   Distributed training setup.  
   Several GPU machines are allocated for each task, and gradients from each task are synchronized and aggregated with separate RMSProp optimizers.
   }
   
\label{fig:net_training}
\end{figure}

\subsection{Self-supervised network architecture implementation details}
This section provides more details on the ``heads'' used in our self-supervised tasks.
The bulk of the changes relative to the original methods (that used shallower networks) involve replacing simple convolutions with residual units.  
Vanishing gradients can be a problem with networks as deep as ours, and residual networks can help alleviate this problem.  
We did relatively little experimentation with architectures for the heads, due to the high computational cost of restarting training from scratch.

\paragraph{Relative Position:}
Given a batch of patches, we begin by running ResNet-v2-101 at a stride of 8.
Most block 3 convolutions produce outputs at stride 16, so running the network at stride 8 requires using convolutions that are dilated, or ``atrous'', such that each neuron receives input from other neurons that are stride 16 apart in the previous layer.
For further details, see the public implementation of ResNet-v2-101 striding in TF-Slim.  
Our patches are 96-by-96, meaning that we get a trunk feature map which is $12 \times 12 \times 1024$ per patch.
For the head, we apply two more residual units.
The first has an output with 1024 channels, a bottleneck with 128 channels, and a stride of 2; the second has an output size of 512 channels, bottleneck with 128 channels, and stride 2.
This gives us a representation of $3 \times 3 \times 512$ for each patch.
We flatten this representation for each patch, and concatenate the representations for patches that are paired.
We then have 3 ``fully-connected'' residual units (equivalent to a convolutional residual unit where the spatial shape of the input and output is $1 \times 1$).
These are all identical, with input dimensionality and output dimensionality of 3*3*512=4608 and a bottleneck dimensionality of 512.  
The final fully connected layer has dimensionality 8 producing softmax outputs.

\paragraph{Colorization:}
As with relative position, we run the ResNet-v2-101 trunk at stride 8 via dilated convolutions.
Our input images are $256 \times 256$, meaning that we have a $32 \times 32 \times 1024$ feature map.
Obtaining good performance when colorization is combined with other tasks seems to require a large number of parameters in the head.
Hence, we use two standard convolution layers with a ReLU nonlinearity: the first has a $2 \times 2$ kernel and 4096 output channels, and the second has a $1 \times 1$ kernel with 4096 channels.
Both have stride 1.
The final output logits are produced by a 1x1 convolution with stride 1 and 313 output channels.
The head has a total of roughly 35M parameters.
Preliminary experiments with a smaller number of parameters showed that adding colorization degraded performance.
We hypothesize that this is because the network's knowledge of color was pushed down into block 3 when the head was small, and thus the representations at the end of block 3 contained too much information about color.

\paragraph{Exemplar:}
As with relative position, we run the ResNet-v2-101 trunk at stride 8 via dilated convolutions.
We resize our images to $256 \times 256$ and sample patches that are $96 \times 96$.
Thus we have a feature map which is $12 \times 12 \times 1024$.
As with relative position, we apply two residual units, the first with an output with 1024 channels, a bottleneck with 128 channels, and a stride of 2; the second has an output size of 512 channels, bottleneck with 128 channels, and stride 2.
Thus, we have a $3 \times 3 \times 512$-dimensional feature, which is used directly to compute the distances needed for our loss.

\paragraph{Motion Segmentation:}
We reshape all images to $240 \times 320$, to better approximate the aspect ratios that are common in our dataset.
As with relative position, we run the ResNet-v2-101 trunk at stride 8 via dilated convolutions.
We expected that, like colorization, motion segmentation could benefit from a large head.  
Thus, we have two $1 \times 1$ conv layers each with dimension 4096, followed by another $1 \times $1 conv layer which produces a single value, which is treated as a logit and used a per-pixel classification.
Preliminary experiments with smaller heads have shown that such a large head is not necessarily important.

\section{Training the Network}

\label{sec:training}
Training a network with nearly 100 hidden layers requires considerable compute power, so we distribute it across several machines.
As shown in figure~\ref{fig:net_training}, each machine trains the network on a single task.  
Parameters for the ResNet-101 trunk are shared across all replicas.
There are also several task-specific layers, or heads, which are shared only between machines that are working on the same task.
Each worker repeatedly computes losses which are then backpropagated to produce gradients.

Given many workers operating independently, gradients are usually aggregated in one of two ways.  
The first option is asynchronous training, where a centralized parameter server receives gradients from workers, applies the updates immediately, and sends back the up-to-date parameters~\cite{recht2011hogwild,dean2012large}.
We found this approach to be unstable, since gradients may be stale if some machines run slowly.
The other approach is synchronous training, where the parameter server accumulates gradients from all workers, applies the accumulated update while all workers wait, and then sends back identical parameters to all workers~\cite{chen2016revisiting}, preventing stale gradients.  
``Backup workers'' help prevent slow workers from slowing down training.
However, in a multitask setup, some tasks are faster than others.
Thus, slow \textit{tasks} will not only slow down the computation, but their gradients are more likely to be thrown out.  

Hence, we used a hybrid approach: we accumulate gradients from all workers that are working on a single task, and then have the parameter servers apply the aggregated gradients from a single task when ready, without synchronizing with other tasks.  
Our experiments found that this approach resulted in faster learning than either purely synchronous or purely asynchronous training, and in particular, was more stable than asynchronous training.

We also used the RMSProp optimizer, which has been shown to improve convergence in many vision tasks versus stochastic gradient descent.  
RMSProp re-scales the gradients for each parameter such that multiplying the loss by a constant factor does not change how quickly the network learns.
This is a useful property in multi-task learning, since different loss functions may be scaled differently.
Hence, we used a separate RMSProp optimizer for each task.
That is, for each task, we keep separate moving averages of the squared gradients, which are used to scale the task's accumulated updates before applying them to the parameters.

For all experiments, we train on 64 GPUs in parallel, and save
checkpoints every roughly 2.4K GPU (NVIDIA K40) hours.  These checkpoints are then
used as initialization for our evaluation tasks.

\section{Evaluation}

\label{sec:evaluation}

Here we describe the three evaluation tasks that we transfer our representation to: image classification, object
category detection, and pixel-wise depth prediction.

\paragraph{ImageNet with Frozen Weights:}
We add a single linear classification layer (a softmax) to the network at the end of block 3, and train on the full ImageNet training set.
We keep all pre-trained weights frozen during training, so we can evaluate raw features.
We evaluate on the ImageNet validation set.  
The training set is augmented in translation and color, following~\cite{szegedy2016inception}, but during evaluation, we don't use multi-crop or mirroring augmentation.
This evaluation is similar to evaluations used elsewhere (particularly Zhang {\it et al.}~\cite{zhang2016colorful}).
Performing well requires good representation of fine-grained object attributes (to distinguish, for example, breeds of dogs).
We report top-5 recall in all charts (except Table~\ref{tab:prev_results}, which reports top-1 to be consistent with previous works).
For most experiments we use only the output of the final ``unit'' of block~3, and use max pooling to obtain a $3 \times 3 \times 1024$ feature vector, which is flattened and used as the input to the one-layer classifier.
For the lasso experiments, however, we use a weighted combination of the (frozen) features from all block 3 layers, and we learn the weight for each layer, following the structure described in section~\ref{sec:lasso}.

\paragraph{PASCAL VOC 2007 Detection:}
We use Faster-RCNN~\cite{ren2015faster}, which trains a single network base with multiple heads for object proposals, box classification, and box localization.
Performing well requires the network to accurately represent object categories and locations, with penalties for missing parts which might be hard to recognize (e.g., a cat's body is harder to recognize than its head).
We fine-tune all network weights.
For our ImageNet pre-trained ResNet-101 model, we transfer all layers up through block 3 from the pre-trained model into the trunk, and transfer block 4 into the proposal categorization head, as is standard.  
We do the same with our self-supervised network, except that we initialize the proposal categorization head randomly.
Following Doersch et al.~\cite{doersch2015unsupervised}, we use multi-scale data augmentation for all methods, including baselines.
All other settings were left at their defaults.
We train on the VOC 2007 trainval set, and evaluate Mean Average Precision on the VOC 2007 test set.
For the lasso experiments, we feed our lasso combination of block 3 layers into the heads, rather than the final output of block 3.

\paragraph{NYU V2 Depth Prediction:}
Depth prediction measures how well a network represents geometry, and how well that information can be localized to pixel accuracy.
We use a modified version of the architecture proposed in Laina {\it et al.}~\cite{laina2016deeper}.
We use the ``up projection'' operator defined in that work, as well as the reverse Huber loss.
We replaced the ResNet-50 architecture with our ResNet-101 architecture, and feed the block 3 outputs directly into the up-projection layers (block 4 was not used in our setup).
This means we need only 3 levels of up projection, rather than 4.
Our up projection filter sizes were 512, 256, and 128.
As with our PASCAL experiments, we initialize all layers up to block 3 using the weights from our self-supervised pre-training, and fine-tune all weights.
We selected one measure---percent of pixels where relative error is below 1.25---as a representative measure (others available in appendix \ref{appendix:depth}).
Relative error is defined as $\max\left(\frac{d_{gt}}{d_{p}},\frac{d_{p}}{d_{gt}}\right)$, where $d_{gt}$ is groundtruth depth and $d_{p}$ is predicted depth.
For the lasso experiments, we feed our lasso combination of block3 layers into the up projection layers, rather than the final output of block 3.

\section{Results: Comparisons and Combinations}

\label{sec:results}
\paragraph{ImageNet Baseline:} As an ``upper bound'' 
on performance, we train a full ResNet-101 model on ImageNet, which
serves as a point of comparison for all our evaluations.  Note that
just under half of the parameters of this network are in block 4,
which are not pre-trained in our self-supervised experiments 
(they are transferred from the ImageNet network only for the Pascal evaluations). 
We use the
standard learning rate schedule of Szegedy et
al.~\cite{szegedy2016inception} for ImageNet training (multiply the
learning rate by $0.94$ every 2 epochs), but we don't use such a schedule
for our self-supervised tasks.

\subsection{Comparing individual self-supervision tasks}

\begin{table*}

\begin{center}
\begin{tabular}{l|c|c|c|c|c|c}
\hline
Pre-training & \multicolumn{2}{|c|}{ImageNet top1} & ImageNet top5 & \multicolumn{2}{|c|}{PASCAL} & NYU \\ \cline{2-7}
           & Prev.& Ours & Ours & Prev. &  Ours & Ours \\
\hline
Relative Position  & 31.7\cite{zhang2016colorful} & 36.21 & 59.21 &  61.7~\cite{doersch2015unsupervised} & 66.75 & 80.54 \\
\hline
Color &32.6\cite{zhang2016colorful} & 39.62 & 62.48 &46.9\cite{zhang2016colorful} & 65.47 & 76.79 \\
\hline
Exemplar & - & 31.51 & 53.08 & -  & 60.94 & 69.57\\
\hline
Motion Segmentation & - & 27.62 & 48.29 &52.2\cite{pathak2016learning} & 61.13 & 74.24 \\
\hline
INet Labels & 51.0\cite{zhang2016colorful} & 66.82 & 85.10 & 69.9\cite{ren2015faster} & 74.17 & 80.06\\
\hline
\end{tabular}
\end{center}

\caption{Comparison of our implementation with previous results on our evaluation tasks: ImageNet with frozen features (left), and PASCAL VOC 2007 mAP with fine-tuning (middle), and NYU depth (right, not used in previous works). Unlike elsewhere in this paper, ImageNet performance is reported here in terms of top 1 accuracy (versus recall at 5 elsewhere).  Our ImageNet pre-training performance on ImageNet is lower than the performance He et al.~\cite{he2016identity} (78.25) reported for ResNet-101 since we remove block 4.}

\label{tab:prev_results}

\end{table*}

Table~\ref{tab:prev_results} shows the performance of individual tasks for the three evaluation
measures. Compared to 
previously-published results, our performance is significantly higher
in all cases, most likely due
to the additional depth of ResNet (cf.\ AlexNet) and additional training time.  
Note, our ImageNet-trained baseline for Faster-RCNN 
is also above the previously published result using ResNet
(69.9 in~\cite{ren2015faster}
cf.\ 74.2 for ours), mostly due to the addition of multi-scale
augmentation for the training images
following~\cite{doersch2015unsupervised}.

Of the self-supervised pre-training
methods, relative position and colorization are
the top performers, with relative position winning on PASCAL and NYU,
and colorization winning on ImageNet-frozen. Remarkably,
relative position performs on-par with ImageNet pre-training on depth
prediction, and the gap is just 7.5\% mAP on PASCAL.  The only task
where the gap remains large is the ImageNet evaluation itself,
which is not surprising since the ImageNet pre-training and evaluation use
the same labels.
Motion segmentation and exemplar training
are somewhat worse than the others, with exemplar worst on Pascal and NYU, and motion
segmentation worst on ImageNet.

\begin{figure}[t]
\begin{center}

   \includegraphics[width=0.98\linewidth]{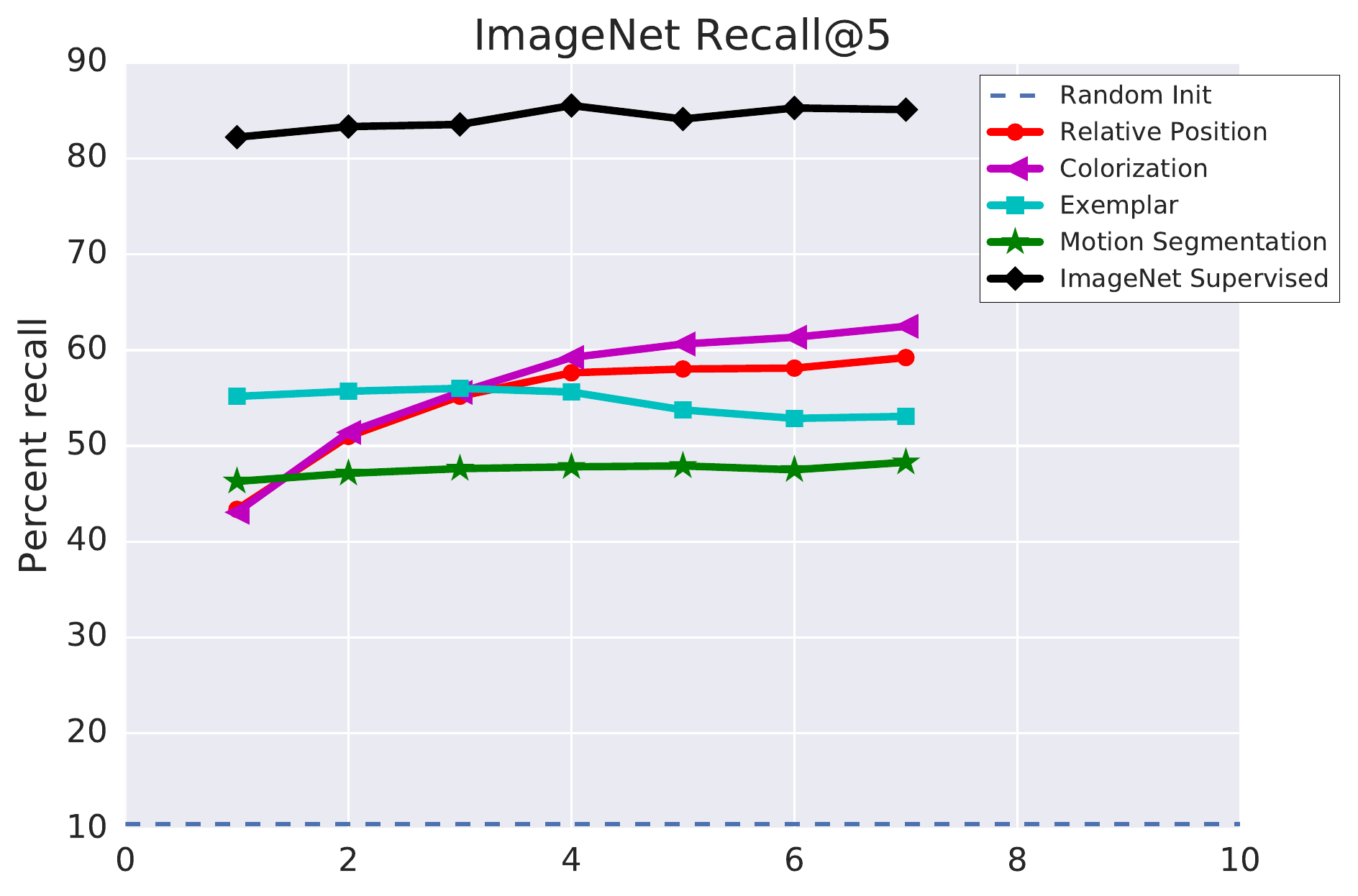}\\
   \includegraphics[width=0.98\linewidth]{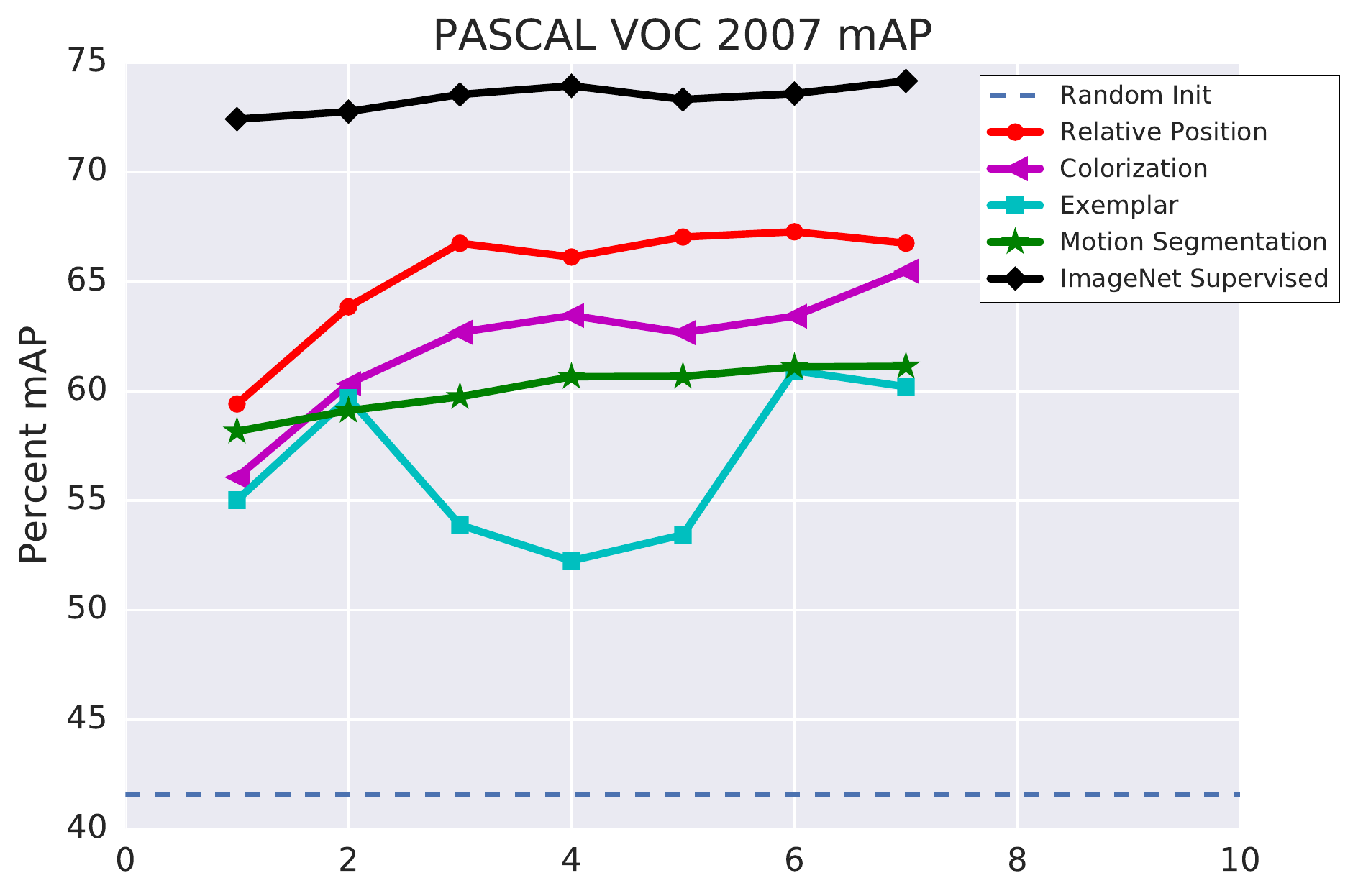}
   \includegraphics[width=0.98\linewidth]{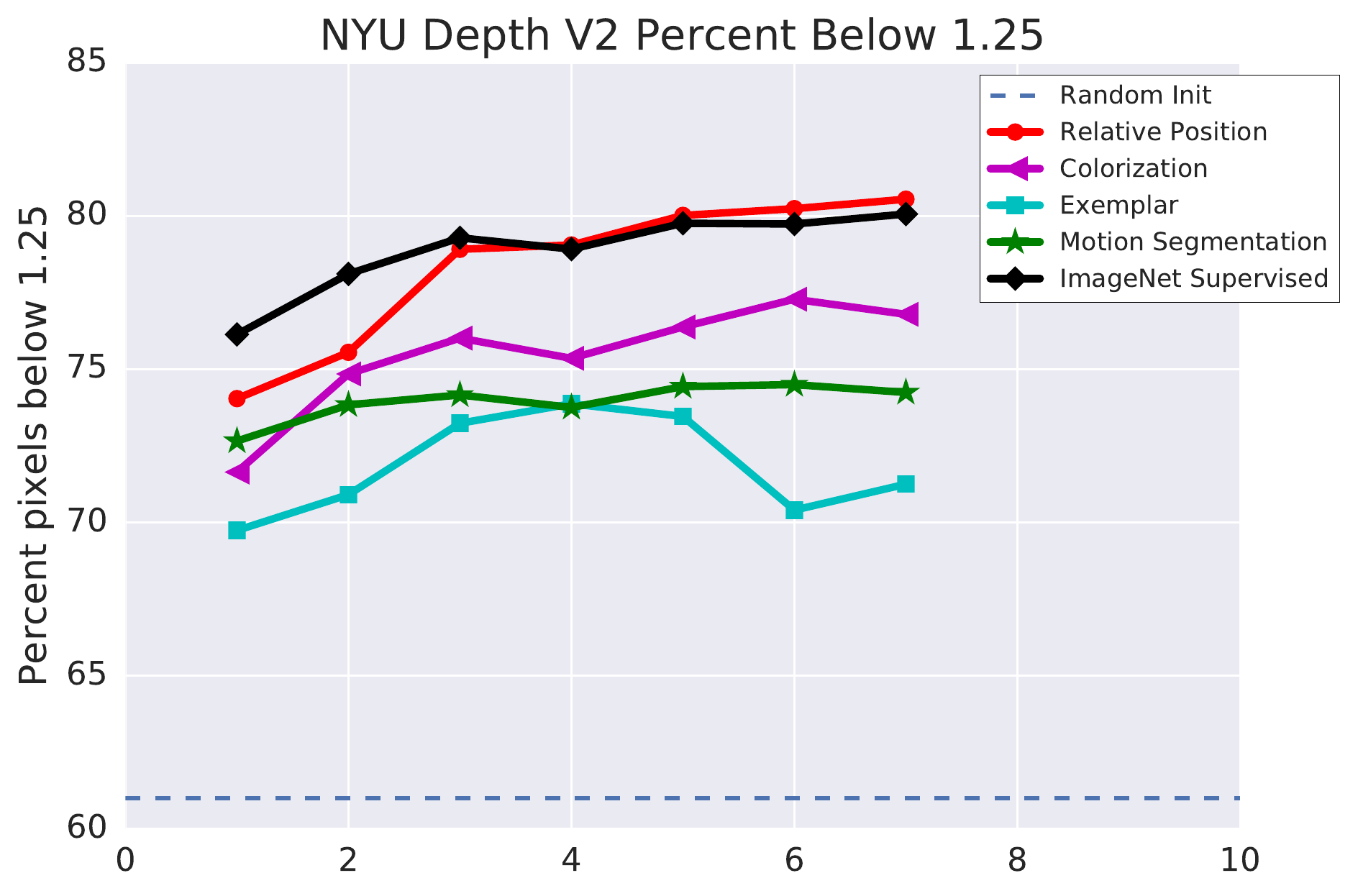}
   
%\fbox{\rule{0pt}{2.5in} \rule{0.9\linewidth}{0pt}}
\end{center}
   \caption{
   Comparison of performance for different self-supervised methods over time.  X-axis is compute time on the self-supervised task ($\sim$2.4K GPU hours per tick).  ``Random Init'' shows performance with no pre-training.
   }
   
\label{fig:base_net_comparison}
\end{figure}

Figure~\ref{fig:base_net_comparison} shows 
how the performance changes as pre-training
time increases (time is on the x-axis).  After 16.8K GPU hours,
performance is plateauing but has not completely saturated, suggesting that
results can be improved slightly given more time.  Interestingly, on
the ImageNet-frozen evaluation, where colorization is winning, the gap
relative to relative position is growing.  Also,
while most algorithms slowly improve performance with training time,
exemplar training doesn't fit this pattern: its performance falls
steadily on ImageNet, and undulates on PASCAL and NYU.  Even
stranger, performance for exemplar is seemingly anti-correlated between Pascal and
NYU from checkpoint to checkpoint.  A possible explanation is that
exemplar training encourages features that aren't invariant beyond the 
training transformations (e.g.\ they aren't invariant to object
deformation or out-of-plane rotation), but are instead sensitive to the
details of textures and low-level shapes.  If
these irrelevant details become prominent in the representation,
they may serve as distractors for the evaluation classifiers.

Note that the random baseline performance is low relative to a shallower network, 
especially the ImageNet-frozen evaluation
(a linear classifier on random AlexNet's conv5 features has top-5
recall of 27.1\%, cf. 10.5\% for ResNet).  
All our pre-trained nets far outperform the random baseline.

The fact that representations learnt by the various self-supervised methods have different
strengths and weaknesses suggests that the features differ.
Therefore, combining methods may yield further improvements.
On the other hand, the lower-performing tasks might drag-down the performance 
of the best ones.  
Resolving this uncertainty is a key motivator for the next section.

\paragraph{Implementation Details:} Unfortunately, intermittent network congestion can slow down
experiments, so we don't measure wall time directly.  Instead, we
estimate compute time for a given task by multiplying the per-task
training step count by a constant factor, which is fixed across all
experiments, representing the average step time when network
congestion is minimal.  We add training cost across all tasks used in an experiment, and
snapshot when the total cost crosses a threshold.  
For relative position, 1 epoch through the ImageNet train set takes roughly 350 GPU hours; for colorization it takes roughly 90 hours; for exemplar nets roughly 60 hours.
For motion segmentation, one epoch through our video dataset takes roughly 400 GPU hours.

\subsection{Na{\"i}ve multi-task combination of self-supervision tasks}
\label{sec:naive}

\begin{table}

\begin{center}
\begin{tabular}{l|c|c|c}
\hline
Pre-training & ImageNet & PASCAL & NYU\\
\hline\hline
RP & 59.21 & 66.75 & \textbf{80.54} \\
\hline
RP+Col  & 66.64 & 68.75 & 79.87\\
\hline
RP+Ex & 65.24 & 69.44 & 78.70\\
\hline
RP+MS & 63.73 & 68.81 & 78.72\\
\hline
RP+Col+Ex & 68.65 & 69.48 & 80.17\\
\hline
RP+Col+Ex+MS & \textbf{69.30} & \textbf{70.53} & 79.25\\
\hline
\hline
INet Labels & 85.10 & 74.17 & 80.06\\
\hline
\end{tabular}
\end{center}

\caption{Comparison of various combinations of self-supervised tasks.  Checkpoints were taken after 16.8K GPU hours, equivalent to checkpoint 7 in Figure~\ref{fig:base_net_comparison}.  Abbreviation key: RP: Relative Position; Col: Colorization; Ex: Exemplar Nets; MS: Motion Segmentation.  Metrics: ImageNet: Recall@5; PASCAL: mAP; NYU: \% Pixels below 1.25.}

\label{tab:combination}
\end{table}

Table~\ref{tab:combination} shows results for combining self-supervised pre-training tasks.
Beginning with one of our strongest performers---relative position---we see that adding any of our other tasks helps performance on ImageNet and Pascal.  
Adding either colorization or exemplar leads to more than 6 points gain on ImageNet.
Furthermore, it seems that the boosts are complementary: adding both colorization and exemplar gives a further 2\% boost.  
Our best-performing method was a combination of all four self-supervised tasks.

To further probe how well our representation localizes objects, we
evaluated the PASCAL detector at a more stringent overlap criterion:
75\% IoU (versus standard VOC 2007 criterion of 50\% IoU).  Our model
gets 43.91\% mAP in this setting, versus the standard ImageNet model's
performance of 44.27\%, a gap of less than half a percent.  Thus, the
self-supervised approach may be especially useful when accurate
localization is important.

\begin{figure}[t]
\begin{center}

   \includegraphics[width=0.98\linewidth]{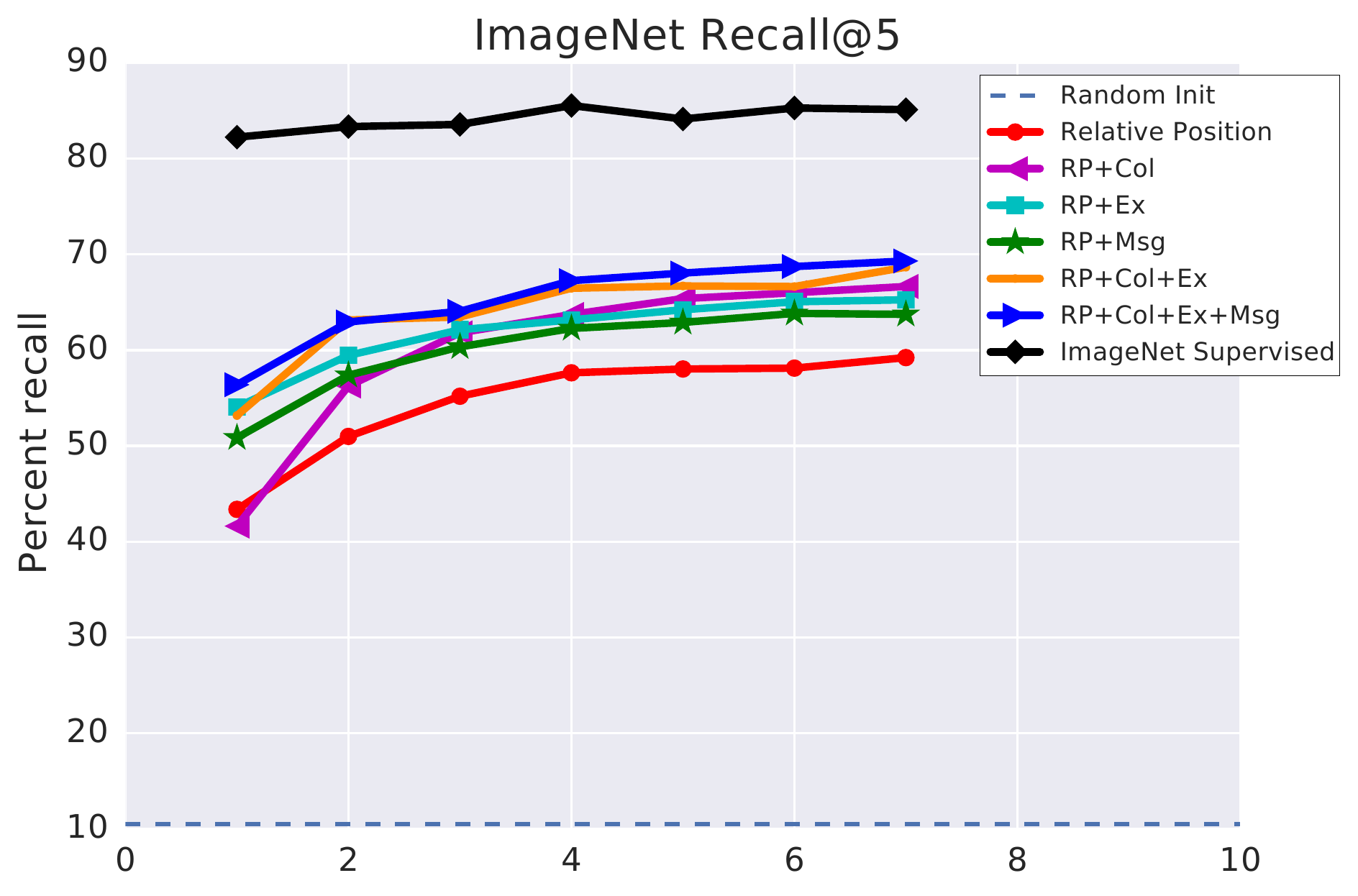}\\
   \includegraphics[width=0.98\linewidth]{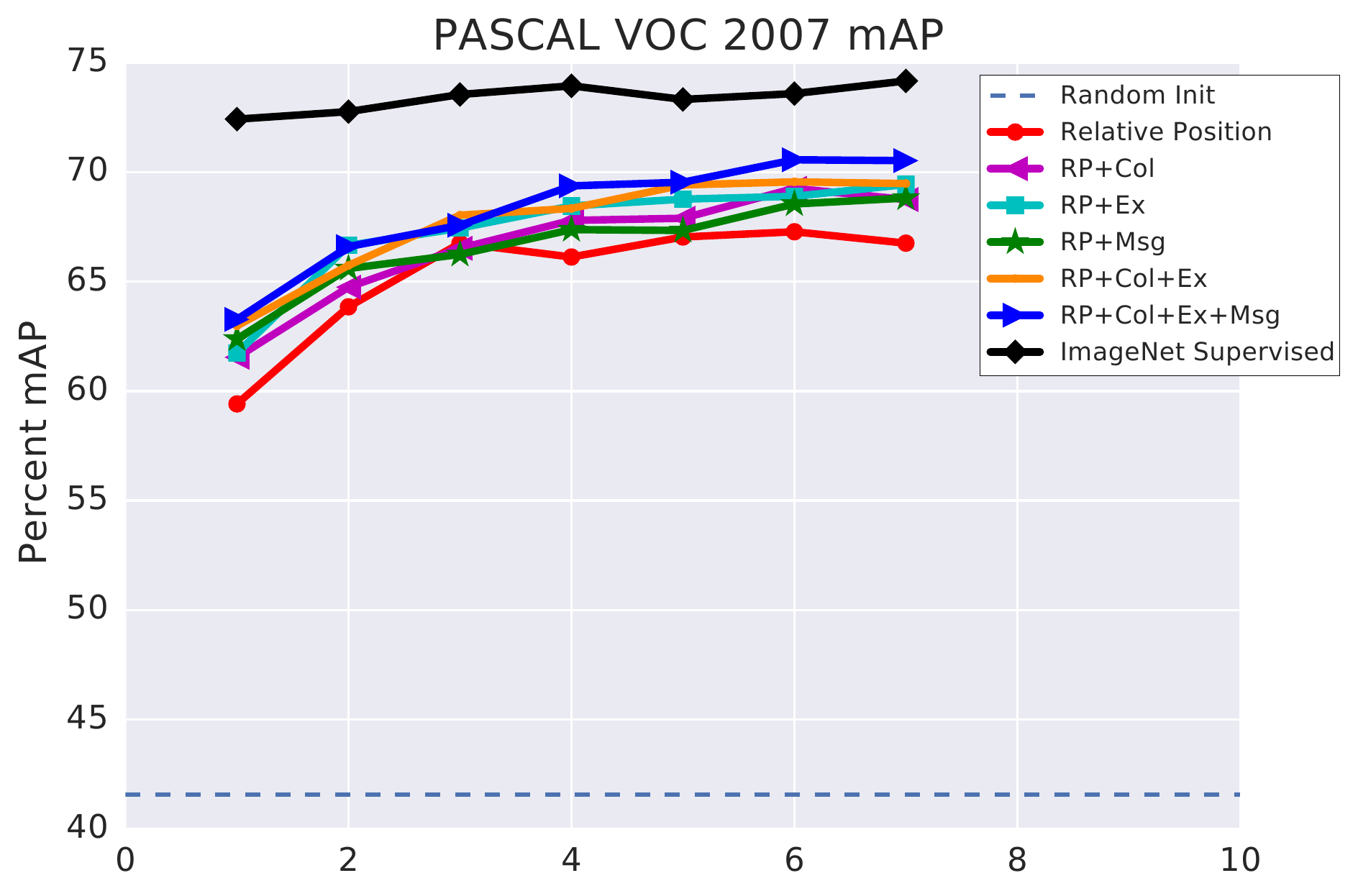}
   \includegraphics[width=0.98\linewidth]{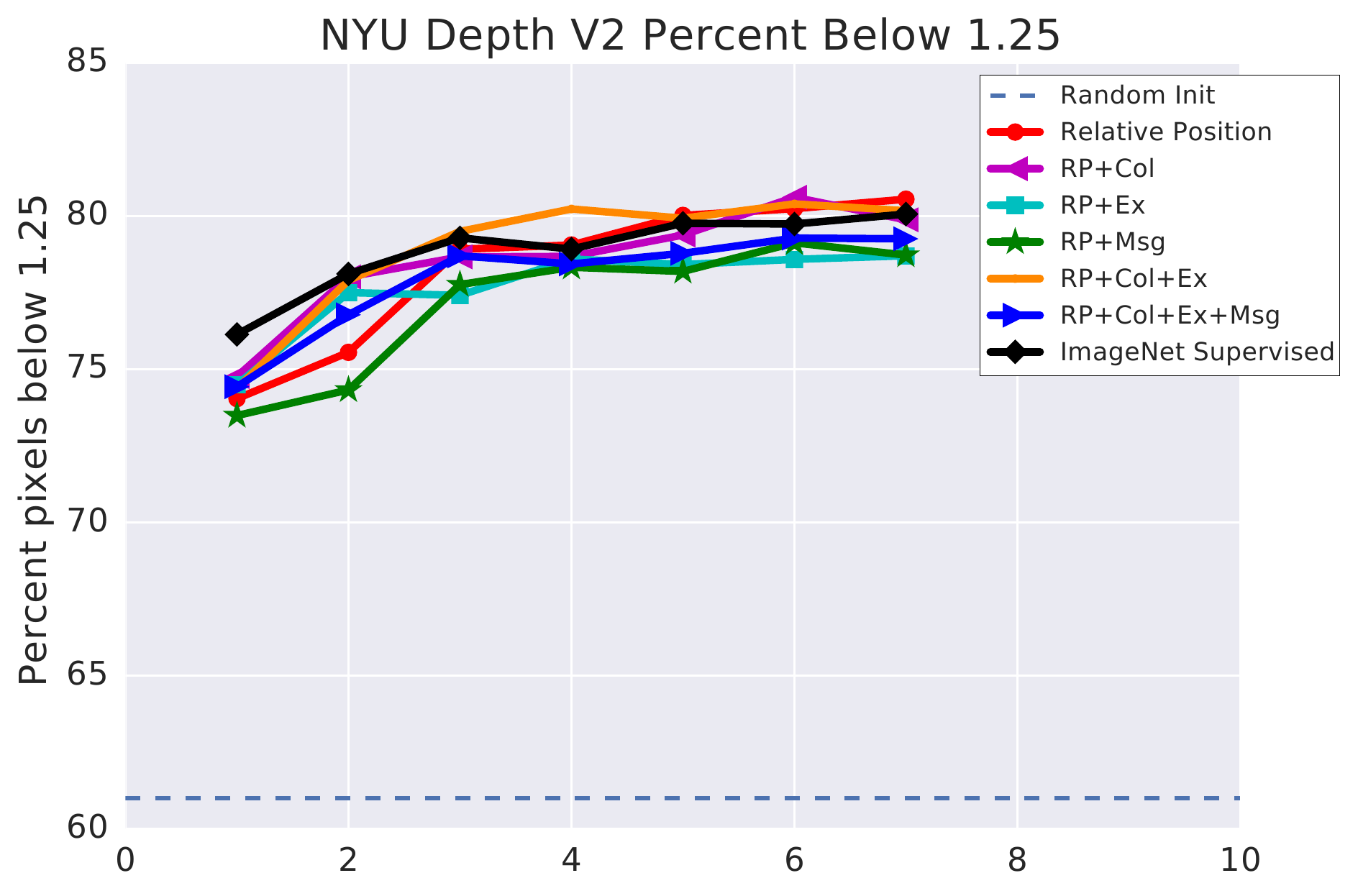}
   
\end{center}
   \caption{
   Comparison of performance for different multi-task self-supervised methods over time.  X-axis is compute time on the self-supervised task ($\sim$2.4K GPU hours per tick).  ``Random Init'' shows performance with no pre-training.
   }
   
\label{fig:multitask_comparison}
\end{figure}

The depth evaluation performance shows far less variation over the single and combinations tasks than the other evaluations.
All methods are on par with ImageNet pre-training, with relative position exceeding this value slightly, and the
combination with exemplar or motion segmentation leading to a slight drop.
Combining relative position with with either exemplar or motion segmentation leads to a considerable improvement over those tasks alone.

Finally, figure~\ref{fig:multitask_comparison} shows how the performance of these methods improves with more training.  
One might expect that more tasks would result in slower training, since more must be learned.
Surprisingly, however the combination of all four tasks performs the best or nearly the best even at our earliest checkpoint.

\subsection{Mediated combination of self-supervision tasks}

\paragraph{Harmonization:} We train two versions of a network on relative position and colorization: one using harmonization to make the relative position inputs look more like colorization, and one without it (equivalent to RP+Col in section~\ref{sec:naive} above).  
As a baseline, we make the same modification to a network trained only on relative position alone: i.e., we convert its inputs to grayscale.
In this baseline, we don't expect any performance boost over the original relative position task, because there are no other tasks to harmonize with.
Results are shown in Table~\ref{tab:harmonization_perf}.
However, on the ImageNet evaluation there is an improvement 
 when we
pre-train using only relative position (due to the change from adding noise to the other two channels
to using grayscale input (three equal channels)), and this improvement follows through to the
the combined relative position and colorization tasks.  The other two evaluation tasks do not show any
improvement with harmonization.
This suggests that our networks are actually quite good at dealing with stark differences between pre-training data domains when the features are fine-tuned at test time.

\begin{table}

\begin{center}
\begin{tabular}{l|c|c|c}
\hline
Pre-training & ImageNet & PASCAL & NYU\\
\hline\hline
RP & 59.21 & 66.75 & 80.23 \\
\hline

RP / H & 62.33 & 66.15 & 80.39\\
\hline
RP+Col & 66.64 & 68.75 &  79.87\\
\hline
RP+Col / H & 68.08 & 68.26 & 79.69\\
\hline
\end{tabular}
\end{center}

\caption{Comparison of methods with and without harmonization, where relative position training is converted to grayscale to mimic the inputs to the colorization network.  H denotes an experiment done with harmonization.  
}

\label{tab:harmonization_perf}
\end{table}

\begin{figure}[t]
\begin{center}

   \includegraphics[width=0.98\linewidth]{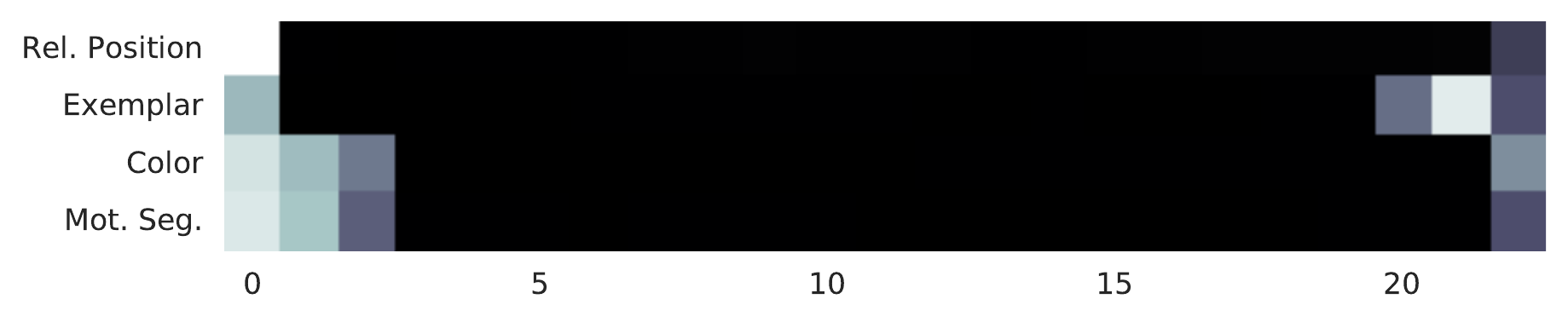}\\
   \includegraphics[width=0.98\linewidth]{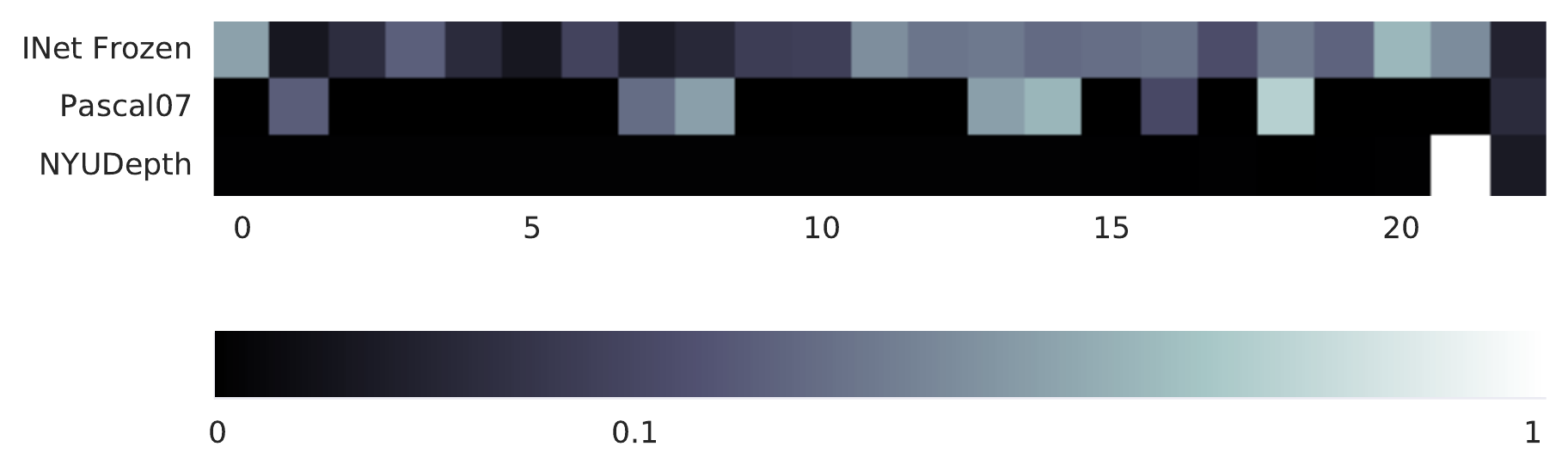}
\end{center}

   \caption{
Weights learned via the lasso technique.  Each row
shows one task: self-supervised tasks on top, evaluation tasks on bottom.  Each square shows $|\alpha|$
for one ResNet ``Unit'' (shallowest layers at the
left).  Whiter colors indicate higher $|\alpha|$, with a nonlinear scale
to make smaller nonzero values easily visible.}

\label{fig:lasso_alpha}
\end{figure}

\begin{table}

\begin{center}
\begin{tabular}{l|c|c|c}
\hline
Net structure & ImageNet & PASCAL & NYU\\
\hline\hline
No Lasso & 69.30 & \textbf{70.53} & 79.25\\
\hline
Eval Only Lasso & \textbf{70.18} & 68.86 & 79.41\\
\hline
Pre-train Only Lasso & 68.09 & 68.49 & 78.96\\
\hline
Pre-train \& Eval Lasso & 69.44 & 68.98 & 79.45\\
\hline
\end{tabular}
\end{center}

\caption{Comparison of performance with and without the lasso technique for factorizing representations, for a network trained on all four self-supervised tasks for 16.8K GPU-hours.  ``No Lasso'' is equivalent to table~\ref{tab:combination}'s RP+Col+Ex+MS.  ``Eval Only'' uses the same pre-trained network, with lasso used only on the evaluation task, while ``Pre-train Only'' uses it only during pre-training.  The final row uses lasso always.  
}

\label{tab:lasso_perf}
\end{table}

\paragraph{Lasso training:} As a first sanity check, Figure~\ref{fig:lasso_alpha} plots the $\alpha$ matrix learned using all four self-supervised tasks.  
Different tasks do indeed select different layers.
Somewhat surprisingly, however, there are strong correlations between the selected layers: most tasks want a combination of low-level information and high-level, semantic information.
The depth evaluation network selects
relatively high-level information, but evaluating on ImageNet-frozen
and PASCAL makes the network select information from several levels, often not the ones that the pre-training tasks use.  
This suggests that, although there are useful features in the learned representation, the final output space for the representation is still losing some information that's useful for evaluation tasks, suggesting a possible area for future work.

The final
performance of this network is shown in Table~\ref{tab:lasso_perf}. There 
are four cases: no lasso, lasso only on the evaluation tasks, lasso only at pre-training time, and lasso
in both self-supervised training and evaluation.
Unsurprisingly, using lasso only for pre-training performs poorly since not all information reaches the final layer.
Surprisingly, however, using the lasso both for self-supervised training and evaluation is not very effective, contrary to previous results advocating that features should be selected from multiple layers for task transfer~\cite{misra2016cross,hariharan2015hypercolumns,rusu2016progressive}.
Perhaps the multi-task nature of our pre-training forces more information to propagate through the entire network, so explicitly extracting information from lower layers is unnecessary.

\section{Summary and extensions}

\label{sec:conclusion}
In this work, our main findings are: (i) 
Deeper networks improve self-supervision over shallow networks; (ii)
Combining self-supervision tasks always improves performance over the tasks alone; (iii)
The gap between ImageNet pre-trained and self-supervision pre-trained with four tasks is nearly 
closed for the VOC detection evaluation, and completely closed for NYU depth, (iv)
Harmonization and lasso weightings only have minimal effects; and, finally,
(v) Combining self-supervised tasks leads to faster training.

There are many opportunities for further improvements: we can add
augmentation (as in the exemplar task) to all tasks; we could add more
self-supervision tasks (indeed new ones have appeared during the
preparation of this paper, e.g.\ \cite{fernando2016self});
we could add further evaluation tasks -- indeed 
depth prediction was not very informative, and replacing it by an alternative shape measurement task such as 
 surface normal prediction may be more reliable;
and we can
experiment with methods for dynamically weighting the importance of
tasks in the optimization. 

It would also be interesting to repeat
these experiments with a deep network such as VGG-16 where
consecutive layers are less correlated,
or with even deeper networks
(ResNet-152, DenseNet~\cite{huang2016densely} and beyond) to tease out
the match between self-supervision tasks and network depth. For the lasso, it might
be worth investigating block level weightings using a group sparsity regularizer.

For the future, given the performance improvements demonstrated in this paper, there is a 
possibility that self-supervision will eventually augment or replace
fully supervised pre-training.

\begin{small}
\paragraph{Acknowledgements:}
Thanks to Relja Arandjelovi\'c, Jo\~{a}o Carreira, Viorica P\u{a}tr\u{a}ucean and Karen Simonyan for helpful discussions.
\end{small}

\appendix
\section{Additional metrics for depth prediction}
\label{appendix:depth}
Previous literature on depth prediction has established several measures of accuracy, since different errors may be more costly in different contexts.  
The measure used in the main paper was percent of pixels where relative depth---i.e., $\max\left(\frac{d_{gt}}{d_{p}},\frac{d_{p}}{d_{gt}}\right)$---is less than 1.25.
This measures how often the estimated depth is very close to being correct.  
It is also standard to measure more relaxed thresholds of relative depth: $1.25^2$ and $1.25^3$.
Furthermore, we can measure average errors across all pixels.
Mean Absolute Error is the mean squared difference between ground truth and predicted values.
Unlike the previous metrics, with Mean Absolute Error the worst predictions receive the highest penalties.
Mean Relative Error weights the prediction error by the inverse of ground truth depth.  
Thus, errors on nearby parts of the scene are penalized more, which may be more relevant for, e.g., robot navigation.

Tables~\ref{tab:methods_alone},~\ref{tab:combinations},~\ref{tab:harmonization}, and~\ref{tab:lasso} are extended versions of tables\ref{tab:prev_results},~\ref{tab:combination},~\ref{tab:harmonization_perf},~\ref{tab:lasso_perf}, respectively.
For the most part, the additional measures tell the same story as the measure for depth reported in the main paper.  
Different self-supervised signals seem to perform similarly relative to one another: exemplar and relative position work best; color and motion segmentation work worse (table~\ref{tab:methods_alone}).
Combinations still perform as well as the best method alone (table~\ref{tab:combinations}).  
Finally, it remains uncertain whether harmonization or the lasso technique provide a boost on depth prediction (tables~\ref{tab:harmonization} and~\ref{tab:lasso}).

\begin{table*}
\begin{center}
\begin{tabular}{l|c|c|c|c|c}
\hline
Evaluation & \multicolumn{3}{|c|}{Higher Better} & \multicolumn{2}{|c}{Lower Better} \\ \cline{2-6}
           & Pct. $< 1.25$ & Pct. $< 1.25^2$& Pct. $< 1.25^3$ & Mean Absolute Error & Mean Relative Error \\
           
\hline\hline
Rel. Pos.  & 80.55 &    94.65 &    98.26 &    0.399 &    0.146 \\
\hline
Color &76.79 &    93.52 &    97.74 &    0.444 &    0.164 \\
\hline
Exemplar & 71.25 &    90.63 &    96.54 &    0.513 &    0.191\\
\hline
Mot. Seg. &74.24 &    92.42 &    97.43 &    0.473 &    0.177\\
\hline
INet Labels & 80.06 &    94.87 &    98.45 &    0.403 &    0.146  \\
\hline
Random & 61.00 &    85.45 &    94.67 &    0.621 &    0.227  \\
\hline
\end{tabular}
\end{center}
\caption{Comparison of self-supervised methods on NYUDv2 depth prediction.  Pct. $< 1.25$ is the same as reported in the paper (Percent of pixels where relative depth---$\max\left(\frac{d_{gt}}{d_{p}},\frac{d_{p}}{d_{gt}}\right)$---is less than 1.25); we give the same value for two other, more relaxed thresholds.  
We also report mean absolute error, which is the simple per-pixel average error in depth, and relative error, where the error at each pixel is divided by the ground-truth depth.}
\label{tab:methods_alone}
\end{table*}

\begin{table*}
\begin{center}
\begin{tabular}{l|c|c|c|c|c}
\hline
Evaluation & \multicolumn{3}{|c|}{Higher Better} & \multicolumn{2}{|c}{Lower Better} \\ \cline{2-6}
           & Pct. $< 1.25$ & Pct. $< 1.25^2$& Pct. $< 1.25^3$ & Mean Absolute Error & Mean Relative Error \\
           
\hline\hline
RP & 80.55 &    94.65 &    98.26 &    0.399 &    0.146 \\
\hline
RP+Col  &79.88 &    94.45 &    98.15 &    0.411 &    0.148 \\
\hline
RP+Ex & 78.70 &    94.06 &    98.13 &    0.419 &    0.151 \\
\hline
RP+MS & 78.72 &    94.13 &    98.08 &    0.423 &    0.153\\
\hline
RP+Col+Ex & 80.17 &    94.74 &    98.27 &    0.401 &    0.149\\
\hline
RP+Col+Ex+MS & 79.26 &    94.19 &    98.07 &    0.422 &    0.152\\
\hline
\end{tabular}
\end{center}
\caption{Additional measures of depth prediction accuracy on NYUDv2 for the na{\"i}ve method of combining different sources of supervision, extending table \ref{tab:combination}.}
\label{tab:combinations}
\end{table*}

\begin{table*}
\begin{center}
\begin{tabular}{l|c|c|c|c|c}
\hline
Evaluation & \multicolumn{3}{|c|}{Higher Better} & \multicolumn{2}{|c}{Lower Better} \\ \cline{2-6}
           & Pct. $< 1.25$ & Pct. $< 1.25^2$& Pct. $< 1.25^3$ & Mean Absolute Error & Mean Relative Error \\
           
\hline\hline
RP & 80.55 &    94.65 &    98.26 &    0.399 &    0.146\\
\hline
RP / H & 80.39 &    94.67 &    98.31 &    0.400 &    0.147\\
\hline
RP+Col & 79.88 &    94.45 &    98.15 &    0.411 &    0.148 \\
\hline
RP+Col / H & 79.69 &    94.28 &    98.09 &    0.411 &    0.152\\

\hline
\end{tabular}
\end{center}
\caption{Additional measures of depth prediction accuracy on NYUDv2 for the harmonization experiments, extending table\ref{tab:harmonization_perf}.}
\label{tab:harmonization}
\end{table*}

\begin{table*}
\begin{center}
\begin{tabular}{l|c|c|c|c|c}
\hline
Evaluation & \multicolumn{3}{|c|}{Higher Better} & \multicolumn{2}{|c}{Lower Better} \\ \cline{2-6}
           & Pct. $< 1.25$ & Pct. $< 1.25^2$& Pct. $< 1.25^3$ & Mean Absolute Error & Mean Relative Error \\
           
\hline\hline
No Lasso &  79.26 &    94.19 &    98.07 &    0.422 &    0.152\\
\hline
Eval Only Lasso & 79.41 &    94.18 &    98.07 &    0.418 &    0.152\\
\hline
Pre-train Only Lasso &78.96 &    94.05 &    97.83 &    0.423 &    0.153\\
\hline
Lasso & 79.45 &    94.49 &    98.26 &    0.411 &    0.151\\
\hline
\end{tabular}
\end{center}
\caption{Additional measures of depth prediction accuracy on NYUDv2 for the lasso experiments, extending table~\ref{tab:lasso_perf}.}
\label{tab:lasso}
\end{table*}

{\small
\bibliographystyle{ieee}
\bibliography{egbib}
}

\end{document}